\pdfoutput=1

\documentclass[11pt]{article}

\usepackage[final]{acl}

\usepackage{times}
\usepackage{latexsym}
\usepackage{amsmath}
\usepackage{amssymb}
\usepackage{bm}
\usepackage{bbm}
\usepackage[ruled,vlined,linesnumbered]{algorithm2e}
\usepackage{setspace}                                
\usepackage{graphicx}   
\usepackage{multirow, array}
\usepackage{booktabs}
\usepackage{color}
\usepackage{arydshln}
\usepackage{tcolorbox}
\tcbuselibrary{skins} 
\usepackage{colortbl}
\usepackage{xcolor}
\usepackage{pifont}
\definecolor{RedOrange}{RGB}{255,69,0}
\definecolor{BlueGreen}{RGB}{0,128,128}
\definecolor{bittersweet}{rgb}{1.0, 0.44, 0.37}
\definecolor{mygreen}{rgb}{0.29, 0.7, 0.48}
\usepackage{pifont}

\definecolor{demphcolor}{RGB}{144,144,144}

\definecolor{mygray}{gray}{0.4}
\definecolor{autopurple}{HTML}{7030A0}
\definecolor{dyna_yellow}{HTML}{BF9000}
\definecolor{adaptive_blue}{HTML}{0070C0}
\definecolor{darksalmon}{rgb}{0.91, 0.59, 0.48}
\definecolor{emerald}{rgb}{0.31, 0.78, 0.47}
\definecolor{green(pigment)}{rgb}{0.0, 0.65, 0.31}
\definecolor{amaranth}{rgb}{0.9, 0.17, 0.31}
\definecolor{iris}{rgb}{0.35, 0.31, 0.81}
\definecolor{uu}{rgb}{0.95, 0.51, 0.51}
\definecolor{spirodiscoball}{rgb}{0.06, 0.75, 0.99}
\definecolor{myforestgreen}{RGB}{34,139,34}

\definecolor{softblue}{RGB}{121,151,161}
\definecolor{softgreen}{RGB}{106,176,106}
\definecolor{softpink}{RGB}{179,127,135}
\definecolor{softpurple}{RGB}{151,134,151}
\definecolor{softbrown}{RGB}{147,126,98}

\newcommand{\blue}[1]{$_{\color{BlueGreen}\downarrow #1}$}
\newcommand{\red}[1]{$_{\color{RedOrange}\uparrow #1}$}

\usepackage{cleveref}
\SetKwInOut{Input}{Input}\SetKwInOut{Output}{Output}

\SetCommentSty{mycommfont}
\SetKwComment{Comment}{$\triangleright$\ }{}

\usepackage{xspace}

\definecolor{ada_blue}{rgb}{0,205,205}
\definecolor{glt_red}{rgb}{109,205,255}
\definecolor{MorandiBlue}{RGB}{118,134,146}

\definecolor{demphcolor}{RGB}{144,144,144}
\definecolor{mygray}{gray}{0.4}
\definecolor{autopurple}{HTML}{7030A0}
\definecolor{dyna_yellow}{HTML}{BF9000}
\definecolor{adaptive_blue}{HTML}{0070C0}
\definecolor{darkgrey}{RGB}{120,120,120}
\definecolor{mygrey}{RGB}{200,200,200}

\newcommand{\XSolidBrush}{\ding{55}}
\newcommand{\Checkmark}{\ding{51}}

\usepackage[font=small,labelfont=bf]{caption} 
\usepackage{makecell}
\usepackage{tabulary}

\definecolor{myblue}{HTML}{00CDCD}
\definecolor{champagne}{rgb}{0.74, 0.83, 0.9}
\definecolor{champagne}{rgb}{0.97, 0.91, 0.81}

\usepackage[T1]{fontenc}

\usepackage[utf8]{inputenc}

 \usepackage[nopatch=footnote]{microtype}

\usepackage{inconsolata}

\usepackage{graphicx}

\title{AgentDropout: Dynamic Agent Elimination for Token-Efficient and High-Performance LLM-Based Multi-Agent Collaboration}

\author{Zhexuan Wang$^{1}$~~
        Yutong Wang$^{1}$~~
        Xuebo Liu$^{1}$\thanks{~~Corresponding Author}~~ 
        \textbf{Liang Ding}$^{2}$~~ \\
        \textbf{Miao Zhang}$^{3}$~~
        \textbf{Jie Liu}$^{4}$~~
        \textbf{Min Zhang}$^{1}$ \\
        $^{1}$Institute of Computing and Intelligence, Harbin Institute of Technology, Shenzhen, China \\
        $^{2}$The University of Sydney, Sydney, Australia \\
        $^{3}$School of Computer Science and Technology, Harbin Institute of Technology, Shenzhen, China \\
        $^{4}$State Key Lab of Smart Farm Technologies and Systems, Harbin Institute of Technology, Harbin, China \\
        \texttt{\{wangzhexuan,wangyutong\}@stu.hit.edu.cn, liangding.liam@gmail.com} \\
        \texttt{\{liuxuebo,zhangmiao,jieliu,zhangmin2021\}@hit.edu.cn}
  }

\begin{document}
\maketitle
\begin{abstract}
Multi-agent systems (MAS) based on large language models (LLMs) have demonstrated significant potential in collaborative problem-solving.
However, they still face substantial challenges of low communication efficiency and suboptimal task performance, making the careful design of the agents' communication topologies particularly important.
Inspired by the management theory that roles in an efficient team are often dynamically adjusted, we propose \textbf{AgentDropout}, which identifies redundant agents and communication across different communication rounds by optimizing the adjacency matrices of the communication graphs and eliminates them to enhance both token efficiency and task performance.
Compared to state-of-the-art methods, AgentDropout achieves an average reduction of 21.6\% in prompt token consumption and 18.4\% in completion token consumption, along with a performance improvement of 1.14 on the tasks.
Furthermore, the extended experiments demonstrate that AgentDropout achieves notable domain transferability and structure robustness, revealing its reliability and effectiveness.
We release our code at \url{https://github.com/wangzx1219/AgentDropout}.
\end{abstract}

\section{Introduction}
In recent years, LLM-based \cite{achiam2023gpt,team2024gemini} agent technologies have achieved robust performance across a wide range of tasks. Prior works have discovered that LLM-based agents~\citep{liang-etal-2024-encouraging,wang2025mixtureofagents, du2024improving, wang-etal-2024-taste, zhuge2024gptswarm,wang2024survey} can emulate human-like collaboration through mutual communication, leading to enhanced performance in tasks and forming MAS that resemble cooperative human teams.

\begin{figure}[t]
\centering 
\includegraphics[height=7.5cm]{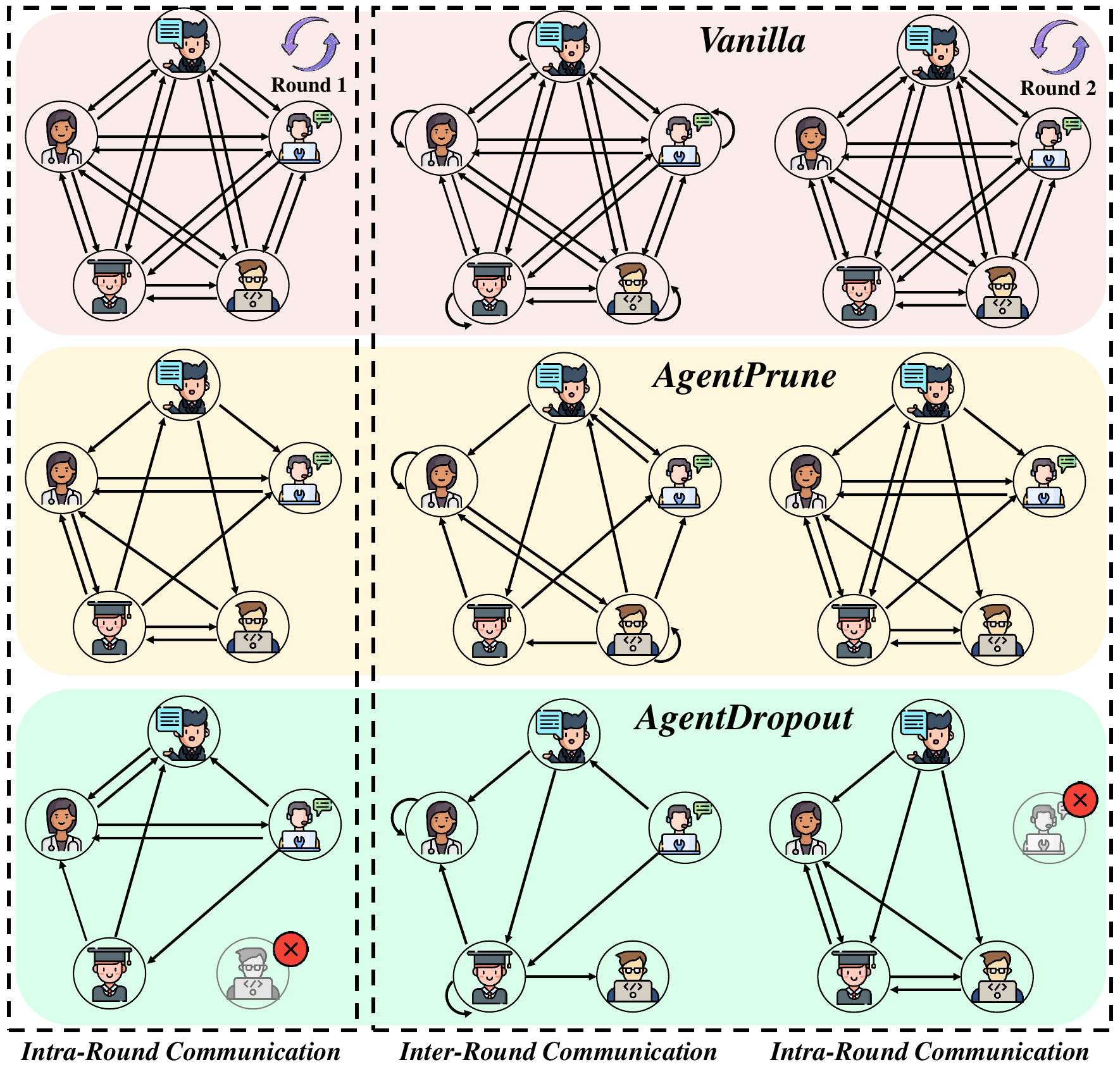}
\caption{An overview of AgentDropout, in comparison with vanilla MAS and AgentPrune~\citep{zhang2025cut}, highlighting its dynamic adjustments of participant roles in each discussion to enhance cooperation effectiveness and efficiency.}
\label{firstpage}
\end{figure}

However, MAS often suffers from high token consumption and inefficiency~\citep{li-etal-2024-improving-multi}. This is primarily due to the frequent generation and communication among multiple agents, which incurs substantial token overhead. Existing methods mainly apply simple agent profile evolution~\citep{chen2024agentverse} or memory evolution~\citep{wang2025delta, qian-etal-2024-experiential,qian2024iterative,gao2024360}, which fails to address the core problem of multi-agent interaction optimization. To address this issue,~\citet{zhuge2024gptswarm} proposed the pruning of inter-graph communication edges, while~\citet{zhang2025cut} defined communication redundancy and introduced a pruning method AgentPrune within MAS. Specifically, existing methods employ trainable communication graph masks to iteratively identify critical message exchanges and prune redundant edges, thereby generating a sparser and more efficient communication graph. However, constrained by the unchanged roles participating in the discussion in each round in MAS, the existing method applies a uniform pruning strategy across multiple communication rounds, which may limit its effectiveness in improving system efficiency and performance.

Drawing an analogy between MAS and human teams, it is evident that dynamic role allocation and team collaboration are pivotal factors in enhancing decision-making efficiency and overall performance in high-performing teams.~\citet{locke2011handbook} have pointed out that an effective team must adaptively adjust the roles and responsibilities of its members based on task requirements,~\citet{kozlowski2006enhancing} further emphasizing the critical role of such dynamic adjustments in improving team decision-making efficacy. Inspired by these insights, we propose AgentDropout, as shown in figure~\ref{firstpage}.
Specifically, AgentDropout iteratively updates the communication topology using a small amount of data, simulating the behavior of dynamic role assignment during multi-round discussions. This process consists of two steps: (1) \textbf{Node Dropout}: Similar to how team members assume different responsibilities at various stages, we train the edge weights and compute the degree of nodes in this step. By selectively retaining or removing certain nodes across different rounds, the flexibility and efficiency of the entire system are enhanced. (2) \textbf{Edge Dropout}: We optimize the communication and collaboration mechanisms within teams by training edge weights and pruning redundant or low-contribution edges to refine the connections between nodes. This approach not only reduces the computational complexity but also enhances the effectiveness of information propagation.

Experiments conducted on a variety of tasks, including reasoning, mathematics, and code generation, demonstrate that AgentDropout outperforms different kinds of MAS and AgentPrune in terms of performance and token efficiency. Our main contributions are as follows:
\begin{itemize}
    \item We observe that in MAS with both intra- and inter-round characteristics, involving different roles in distinct rounds of discussion is more efficient and leads to better performance compared to engaging the same roles in every communication round.
    \item We propose AgentDropout, a simple and flexible optimization method for MAS. It adjusts the overall communication topology through two steps: \textbf{Node Dropout} and \textbf{Edge Dropout}.
    \item We observe that for larger-scale and more powerful LLMs, the performance of MAS designed on top of them can still be improved through interaction. AgentDropout can further enhance efficiency, domain transferability, and structural robustness.
\end{itemize}
\section{Background}

\begin{figure*}[t]
\centering 
\includegraphics[height=11cm]{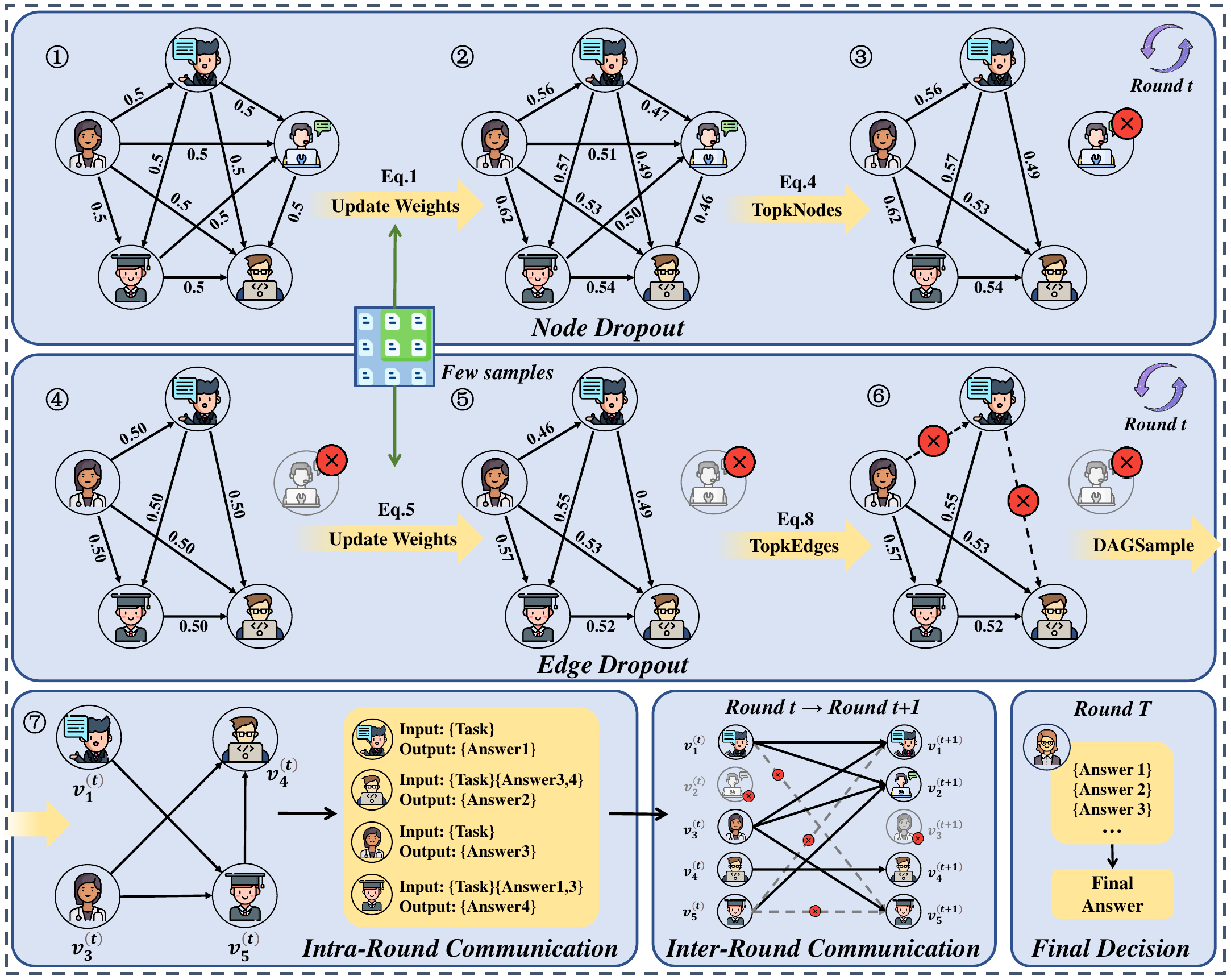}
\caption{The overall process of AgentDropout. The first and second rows present Node Dropout and Edge Dropout procedures, respectively. The third row illustrates the cooperative reasoning process for both intra- and inter-round communication, as well as the generation of the final answer. For simplicity, the edges of inter-round communication are omitted in the demonstration of step 1\textasciitilde7, but they also participate in the update, dropout, and reasoning processes as shown in the middle block of the third row.}
\label{main_fig}
\end{figure*}

\subsection{LLM-based MAS as Graph}
Recently, \citet{chan2024chateval,jiang-etal-2023-llm,wu2023autogen} have proposed representing MAS using directed communication graphs, denoted as $\boldsymbol{G} = (\boldsymbol{V}, \boldsymbol{E}, \boldsymbol{F})$.
$\boldsymbol{V} = \{v_i\}_{i=1}^N$ is a set of nodes, and each node represents an agent.
$\boldsymbol{E} \subset \boldsymbol{V} \times \boldsymbol{V}$ is a set of edges, where each edge $(v_i, v_j) \in \boldsymbol{E} \;(i \neq j)$ denotes the communication path from agent $v_i$ to agent $v_j$.
$\boldsymbol{F} = \{f_i\}_{i=1}^N$ is a set of mapping functions that abstract the agent inference process.
Each node $v_i$ receives a query $q$ and the outputs of its in-neighbors $\boldsymbol{Z}_i = \{o_j \mid (v_j, v_i) \in \boldsymbol{E}\}$ to produce its own output $o_i = f_i(q, \boldsymbol{Z}_i)$.

For some complex tasks, the entire communication graph needs to be run over multiple rounds to obtain the final answer, with each round operating on the communication graph defined as: $\boldsymbol{G}^{(t)}=\left(\boldsymbol{V}^{(t)}, \boldsymbol{E}_\text{intra}^{(t)}, \boldsymbol{F}^{(t)}\right)$, where $\boldsymbol{E}_\text{intra}^{(t)}$ denotes the \textbf{intra-round edges} which only connects nodes in the same round.
Besides, the output of any node in step $t-1$ can be passed to any node in step $t$, forming \textbf{inter-round edges} $\boldsymbol{E}_\text{inter}^{(t)} \subset \boldsymbol{V}^{(t-1)} \times \boldsymbol{V}^{(t)}$ between graph $\boldsymbol{G}^{(t - 1)}$ and $\boldsymbol{G}^{(t)}$.
The subgraphs for all steps $\left\{\boldsymbol{G}^{(1)},\boldsymbol{G}^{(2)},\dots,\boldsymbol{G}^{(T)}\right\}$ are combined using $\boldsymbol{E}_\text{inter}^{(t)}$ to form a multi-round communication graph $\mathcal{\boldsymbol{G}} = (\mathcal{\boldsymbol{V}}, \mathcal{\boldsymbol{E}}, \mathcal{\boldsymbol{F}})$, where $\mathcal{\boldsymbol{V}} = \bigcup_{t} \boldsymbol{V}^{(t)}$, $\mathcal{\boldsymbol{E}} = \left(\bigcup_{t} \boldsymbol{E}_\text{intra}^{(t)}\right) \cup \left(\bigcup_{t} \boldsymbol{E}_\text{inter}^{(t)}\right)$, and $\mathcal{\boldsymbol{F}} = \bigcup_{t} \boldsymbol{F}^{(t)}$.

\subsection{Communication Redundancy in MAS}
MAS often incurs high token consumption associated with the frequent communications between agents.
\citet{zhang2025cut} defined communication redundancy in MAS -- excessive information flow that does not help improve performance.
Considering a subgraph $\mathcal{\boldsymbol{G}}_\text{sub} = (\mathcal{\boldsymbol{V}}, \mathcal{\boldsymbol{E}}', \mathcal{\boldsymbol{F}})$, where $\mathcal{\boldsymbol{E}}' \subseteq \mathcal{\boldsymbol{E}}$,
communication redundancy is considered to exist if
$\mu(\mathcal{\boldsymbol{G}}_\text{sub}) \geq \mu(\mathcal{\boldsymbol{G}})$,
where $\mu(\cdot)$ is the task-specified utility function evaluating the performance of the MAS.
Redundant edges are those in $\mathcal{\boldsymbol{E}} \setminus \mathcal{\boldsymbol{E}}'$.

AgentPrune is an approach to reduce communication redundancy.
By iteratively optimizing two trainable matrix masks (intra- and inter-round), AgentPrune parameterizes and prunes the edges in the communication graph to get rid of redundant edges (i.e. $\mathcal{\boldsymbol{E}} \setminus \mathcal{\boldsymbol{E}}'$).

\section{AgentDropout}

\subsection{Overview}

\paragraph{Motivation}
We argue that communication redundancy in MAS primarily arises from two sources: redundant agent roles and redundant information exchange. 
AgentPrune prunes redundant information exchange but neglects the existence of redundant agent roles and applies the same pruning strategy across all communication rounds, which still leaves redundancy in MAS and may result in performance degradation since different combinations of agent roles might be better suited for different stages of task solving.
To this end, we propose a strategy called \textbf{AgentDropout} to enhance MAS efficiency and performance.
We aim to remove the node that contributes least to the current task from each round's communication graph $\boldsymbol{G}^{(t)}$ and eliminate both intra- and inter-round edges from the global graph $\mathcal{\boldsymbol{G}}$ that have a minimal positive impact on the task solution.

\paragraph{Preparation}
We first convert the communication graph into a weighted graph
$\tilde{\mathcal{\boldsymbol{G}}}$,
where each edge in the graph is a trainable continuous weight in the range of 0 to 1.
Let the adjacency matrix set of $\tilde{\mathcal{\boldsymbol{G}}}$ be
$\tilde{\mathcal{\boldsymbol{A}}} = \tilde{\mathcal{\boldsymbol{A}}}_\text{intra} \cup \tilde{\mathcal{\boldsymbol{A}}}_\text{inter}$, where
$\tilde{\mathcal{\boldsymbol{A}}}_\text{intra} = \bigcup_{t}\tilde{\boldsymbol{A}}_\text{intra}^{(t)}$
is the subset containing intra-round adjacency matrices,
$\tilde{\boldsymbol{A}}_\text{intra}^{(t)} \in [0, 1]^{N \times N}$
is the intra-round adjacency matrix of the round $t$ communication graph and
$\tilde{\mathcal{\boldsymbol{A}}}_\text{inter} = \bigcup_{t}\tilde{\boldsymbol{A}}_\text{inter}^{(t)}$
is the subset containing inter-round adjacency matrices,
$\tilde{\boldsymbol{A}}_\text{inter}^{(t)} \in [0, 1]^{N \times N}$
is the inter-round adjacency matrix between round $t-1$ and $t$.
The actual communication graph used during inference $\mathcal{\boldsymbol{G}}$ is obtained by algorithm \texttt{DAGSample}, where each edge is sampled with its own weight as probability, and the output graph is guaranteed to be a directed acyclic graph (DAG).
AgentDropout consists of two steps to optimize the graph: \textbf{Node Dropout} and \textbf{Edge Dropout}.

\subsection{Node Dropout}
In human teams tackling complex tasks, varying the roles involved in discussions across different rounds can enhance the team's performance by focusing on the most relevant contributors at each stage.
Therefore, we first remove nodes from the communication graph that have less contribution to each round of communication.
As shown in the first row in Figure \ref{main_fig}, we initialize the intra- and inter-round adjacency matrices $\tilde{\mathcal{\boldsymbol{A}}}_\text{intra} = \left\{\tilde{\boldsymbol{A}}_\text{intra}^{(t)}\right\}_t$
and $\tilde{\mathcal{\boldsymbol{A}}}_\text{inter} = \left\{\tilde{\boldsymbol{A}}_\text{inter}^{(t)}\right\}_t$
by setting all the edge weights to $0.5$.\footnote{In detail, in the intra-round adjacency matrices, the diagonal elements are 0, and the other elements are 0.5. The inter-round adjacency matrices are upper triangular matrices with diagonal elements as 0 and others as 0.5.}
We then train $\tilde{\mathcal{\boldsymbol{A}}}_\text{intra}$ to get an optimized weighted communication graph, which is prepared for Node Dropout.
The optimization objective is simply high task performance, formulated as:
\begin{equation}\label{eq:node_dropout_optimization}\small
    \operatorname*{arg\,max}_{\tilde{\mathcal{\boldsymbol{A}}}_\text{intra}} \mathbb{\boldsymbol{E}}_{\mathcal{\boldsymbol{G}}\sim\mathbb{\boldsymbol{G}}}\left[\mu\left(\mathcal{\boldsymbol{G}}\right)\right]
\end{equation}
where $\mathbb{\boldsymbol{G}}$ represents the feasible domain of graph sampling.
Since the utility function $\mu(\cdot)$ often depends on external APIs, it is commonly non-differentiable.
Therefore, we apply a reinforcement algorithm that employs an unbiased policy gradient estimation, as proposed by \citet{zhuge2024gptswarm}.
This approach uses the probability-weighted average performance of a limited number of samples to approximate the overall performance expectation.
We sample $M$ communication graphs upon $\tilde{\mathcal{\boldsymbol{A}}}_\text{intra}, \tilde{\mathcal{\boldsymbol{A}}}_\text{inter}$ using \texttt{DAGSample} and run test instances on each graph $\mathcal{\boldsymbol{G}}_m$ to obtain the performance score $\mu\left(\mathcal{\boldsymbol{G}}_m\right)$:
\begin{equation}\label{eq:expectation_approximate}\small
\begin{aligned}
    \nabla_{\tilde{\mathcal{\boldsymbol{A}}}_\text{intra}}\; &\mathbb{\boldsymbol{E}}_{\mathcal{\boldsymbol{G}}\sim\mathbb{\boldsymbol{G}}}\left[\mu\left(\mathcal{\boldsymbol{G}}\right)\right] \\
    &\approx \frac{1}{M}\sum_{m=1}^M\mu\left(\mathcal{\boldsymbol{G}}_m\right) \nabla_{\tilde{\mathcal{\boldsymbol{A}}}_\text{intra}}\log\left(p_{\tilde{\mathcal{\boldsymbol{A}}}_\text{intra}}\left(\mathcal{\boldsymbol{G}}_m\right)\right)
\end{aligned}
\end{equation}
where $p_{\tilde{\mathcal{\boldsymbol{A}}}_\text{intra}}\left(\mathcal{\boldsymbol{G}}_m\right)$ represents the probability of sampling the communication graph $\mathcal{\boldsymbol{G}}_m = \left(\mathcal{\boldsymbol{F}}_m,\mathcal{\boldsymbol{E}}_{m},\mathcal{\boldsymbol{V}}_m\right)$ given the current weighted intra-round adjacency matrices $\tilde{\mathcal{\boldsymbol{A}}}_\text{intra}$:
\begin{equation}\small
    p_{\tilde{\mathcal{\boldsymbol{A}}}_\text{intra}}\left(\mathcal{\boldsymbol{G}}_m\right) = \prod_{t}{\prod_{\left(v_i,v_j\right)\in\boldsymbol{E}_{m,\text{intra}}^{(t)}}{\tilde{\boldsymbol{A}}_\text{intra}^{(t)}}\left[i,j\right]}
\end{equation}
With the gradient calculated, we update $\tilde{\mathcal{\boldsymbol{A}}}_\text{intra}$ using the gradient ascent method.

After the graph optimization, we select the nodes with the smallest weighted in- and out-degree in each intra-round graph and remove these nodes with all the edges connected to them:
\begin{equation}\label{eq:select_nodes}\small
    \mathcal{\boldsymbol{V}}_\Delta = \bigcup_{t} \; \left(\boldsymbol{V}^{(t)}\setminus\texttt{TopkNodes}\left(\tilde{\boldsymbol{A}}_\text{inter}^{(t)}, 1-\alpha\right)\right)
\end{equation}
where $\texttt{TopkNodes}\left(\cdot\right)$ is a function returning a set of $k$ nodes with the largest sum of in-degree and out-degree according to the adjacency matrix, and $\alpha$ is the node dropout rate.
$\tilde{\mathcal{\boldsymbol{A}}}_\text{intra}$ and $\tilde{\mathcal{\boldsymbol{A}}}_\text{inter}$ is also modified simultaneously by removing all rows and columns associated with the removed nodes.

\begin{table*}[t]
  \centering
  \scalebox{0.75}{
\begin{tabular}{l|cc|ccccccc}
\Xhline{1.2pt}
\rowcolor{CadetBlue!20} 
{\textbf{Method}} & \textbf{Edge DR.} & \textbf{Node DR.} & \textbf{MMLU} & \textbf{GSM8K} & \textbf{AQuA} & \textbf{MultiArith} & \textbf{SVAMP} & \textbf{HumanEval} & {\textbf{Avg.}} \\
\Xhline{1.2pt}
\multicolumn{10}{c}{Base model: Llama3-8B-Instruct} \\
\hline
Vanilla & \textcolor{darksalmon}{\XSolidBrush} & \textcolor{darksalmon}{\XSolidBrush} & 53.59 & 70.23 & 41.67 & 91.11 & 75.00 & 53.33 & 64.16\\

CoT & \textcolor{darksalmon}{\XSolidBrush} & \textcolor{darksalmon}{\XSolidBrush} & 56.86\red{3.27} & 70.47\red{0.24} & 43.75\red{2.08} & 92.25\red{1.14} & 76.17\red{1.67} & 54.17\red{0.84} & 65.61\red{1.45}\\
\hdashline
$\text{MAS}_{round=1}$ & \textcolor{darksalmon}{\XSolidBrush} & \textcolor{darksalmon}{\XSolidBrush} & 56.21\red{2.62} & 69.30\blue{0.97} & 45.29\red{3.62} & 91.67\red{0.55} & 76.67\red{1.67} & 48.33\blue{5.00} & 64.58\red{0.42}\\

$\text{MAS}_{round=T}$ & \textcolor{darksalmon}{\XSolidBrush} &  \textcolor{darksalmon}{\XSolidBrush} & 60.13\red{6.54} & 71.48\red{1.25} & 45.41\red{3.75} & 90.56\blue{0.55} & 77.56\red{2.56} & 49.17\blue{4.16} & 65.72\red{1.56} \\

AgentPrune & \textcolor{green(pigment)}{\Checkmark} & \textcolor{darksalmon}{\XSolidBrush} & 60.78\red{7.19} & 71.02\red{0.79} & 47.22\red{5.55} & 90.00\blue{1.11} & 78.34\red{3.34} & {51.67}\blue{1.66} & 66.51\red{2.35}\\

AgentDropout & \textcolor{green(pigment)}{\Checkmark} & \textcolor{green(pigment)}{\Checkmark} & \textbf{62.75}\red{9.16} & \textbf{73.13}\red{2.90} & \textbf{47.78}\red{6.11} & \textbf{92.59}\red{1.48} & \textbf{80.11}\red{5.11} & \textbf{55.84}\red{2.51} & \textbf{68.70}\red{4.54} \\
\hline
\multicolumn{10}{c}{Base model: Qwen2.5-72B-Instruct} \\
\hline
Vanilla & \textcolor{darksalmon}{\XSolidBrush} & \textcolor{darksalmon}{\XSolidBrush} & 82.35 & 91.02 & 83.75 & 98.89 & 92.67 & 85.28 & 88.99 \\

CoT & \textcolor{darksalmon}{\XSolidBrush} & \textcolor{darksalmon}{\XSolidBrush} & 83.66\red{1.31} & 92.19\red{1.17} & 84.58\red{0.83} & \textbf{100}\red{1.11} & 93.35\red{0.68} & 86.67\red{1.39} & 90.08\red{1.09} \\
\hdashline
$\text{MAS}_{round=1}$ & \textcolor{darksalmon}{\XSolidBrush} & \textcolor{darksalmon}{\XSolidBrush} & 82.35\red{0.00} & 93.52\red{2.50} & 84.58\red{0.83} & \textbf{100}\red{1.11} & 92.36\blue{0.31} & 84.17\blue{1.11} & 89.50\red{0.51} \\

$\text{MAS}_{round=T}$ & \textcolor{darksalmon}{\XSolidBrush} &  \textcolor{darksalmon}{\XSolidBrush} & 84.31\red{1.96} & 93.28\red{2.26} & 85.83\red{2.08} & \textbf{100}\red{1.11} & 94.07\red{1.40} & 87.08\red{1.80} & 90.76\red{1.77} \\

AgentPrune & \textcolor{green(pigment)}{\Checkmark} & \textcolor{darksalmon}{\XSolidBrush} & 83.66\red{1.31} & 93.67\red{2.65} & 87.08\red{3.33} & 99.44\red{0.56} & 94.33\red{1.66} & 86.67\red{1.39} & 90.81\red{1.82} \\

AgentDropout & \textcolor{green(pigment)}{\Checkmark} & \textcolor{green(pigment)}{\Checkmark} & \textbf{84.97}\red{2.62} & \textbf{93.75}\red{2.73} & \textbf{87.50}\red{3.75} & \textbf{100}\red{1.11} & \textbf{95.34}\red{2.67} & \textbf{87.92}\red{2.64} & \textbf{91.58}\red{2.59} \\
\hline
\multicolumn{10}{c}{Base model: Deepseek-V3-671B-Instruct} \\
\hline
Vanilla & \textcolor{darksalmon}{\XSolidBrush} & \textcolor{darksalmon}{\XSolidBrush} & 84.97 & 94.68 & 84.58 & \textbf{100} & 93.67 & 88.43 & 91.06\\

CoT     & \textcolor{darksalmon}{\XSolidBrush} & \textcolor{darksalmon}{\XSolidBrush} & 84.31\blue{0.66} & 95.15\red{0.47} & 85.42\red{0.84} & \textbf{100}\red{0.00} & 93.94\red{0.27} & 89.26\red{0.83} & 91.34\red{0.28}\\
\hdashline
$\text{MAS}_{round=1}$ & \textcolor{darksalmon}{\XSolidBrush} & \textcolor{darksalmon}{\XSolidBrush} & 89.98\red{5.01} & 95.54\red{0.86} & 86.67\red{2.19} & \textbf{100}\red{0.00} & 93.50\blue{0.17} & 89.17\red{0.74} & 92.63\red{1.57} \\

$\text{MAS}_{round=T}$ & \textcolor{darksalmon}{\XSolidBrush} &  \textcolor{darksalmon}{\XSolidBrush} & 89.54\red{4.57} & 95.49\red{0.81} & 87.50\red{2.92} & \textbf{100}\red{0.00} & 94.33\red{0.66} & 89.26\red{0.83} & 92.69\red{1.63} \\

AgentPrune & \textcolor{green(pigment)}{\Checkmark} & \textcolor{darksalmon}{\XSolidBrush} & 90.20\red{5.23} & 95.49\red{0.81} & 87.92\red{3.34} & \textbf{100}\red{0.00} & 95.00\red{1.33} & 90.91\red{2.47} & 93.25\red{2.19} \\

AgentDropout & \textcolor{green(pigment)}{\Checkmark} & \textcolor{green(pigment)}{\Checkmark} & \textbf{90.85}\red{5.88} & \textbf{95.63}\red{0.95} & \textbf{88.33}\red{3.75} & \textbf{100}\red{0.00} & \textbf{95.79}\red{2.12} & \textbf{91.74}\red{3.31} & \textbf{93.72}\red{2.66} \\
\Xhline{1.2pt}
\end{tabular}
} 
  \caption{Performance comparison between AgentDropout and other baseline reasoning techniques. \textbf{Edge DR.} and \textbf{Node DR.} represent the Edge Dropout and Node Dropout methods, respectively.}     
  \label{main_result}  
\end{table*}

\subsection{Edge Dropout}
Promoting more focused and efficient interactions between roles also improves overall teamwork effectiveness of task completion.
Thus, we delete the intra- and inter-round edges in the communication graph that are less important to enhance communication efficiency.
As shown in the second row in Figure \ref{main_fig}, we reinitialize the intra- and inter-round adjacency matrices $\tilde{\boldsymbol{A}}_\text{intra}^{(t)}$, $\tilde{\boldsymbol{A}}_\text{inter}^{(t)}$ after Node Dropout and train \textbf{both} of them from scratch.
Differently, we aim at both high performance and token efficiency this time, formulated as:
\begin{equation}\label{eq:edge_dropout_optimization}\small
\begin{aligned}
    \operatorname*{arg\,max}_{\tilde{\mathcal{\boldsymbol{A}}}_\text{intra},\tilde{\mathcal{\boldsymbol{A}}}_\text{inter}} \;&\mathbb{\boldsymbol{E}}_{\mathcal{\boldsymbol{G}}\sim\mathbb{\boldsymbol{G}}_\text{Node}}\left[\mu\left(\mathcal{\boldsymbol{G}}\right)\right] \\
    & - \left[\sum_{t=1}^{T}\text{rank}\left(\tilde{\boldsymbol{A}}_\text{intra}^{(t)}\right) + \sum_{t=2}^{T}{\text{rank}\left(\tilde{\boldsymbol{A}}_\text{inter}^{(t)}\right)}\right]
\end{aligned}
\end{equation}
where $\mathbb{\boldsymbol{G}}_\text{Node}$ represents the feasible domain of graph sampling after Node Dropout.
We also use the weighted average performance of the samples to approximate the performance expectation as in Equation (\ref{eq:expectation_approximate}).
However, different from Equation (\ref{eq:expectation_approximate}), we calculate the probability and gradient for both $\tilde{\mathcal{\boldsymbol{A}}}_\text{intra}$ and $\tilde{\mathcal{\boldsymbol{A}}}_\text{inter}$ here:
\begin{equation}\small
\begin{aligned}
    p_{\tilde{\mathcal{\boldsymbol{A}}}_\text{intra}, \tilde{\mathcal{\boldsymbol{A}}}_\text{inter}}\left(\mathcal{\boldsymbol{G}}_m\right) = &\prod_{t=1}^T{\prod_{\left(v_i,v_j\right)\in\boldsymbol{E}_{m,\text{intra}}^{(t)}}{\tilde{\boldsymbol{A}}_\text{intra}^{(t)}}\left[i,j\right]} \\
    \cdot &\prod_{t=2}^T{\prod_{\left(v_i,v_j\right)\in\boldsymbol{E}_{m,\text{inter}}^{(t)}}{\tilde{\boldsymbol{A}}_\text{inter}^{(t)}}\left[i,j\right]}
\end{aligned}
\end{equation}

The second term here represents low-rank sparsity, which serves as a proxy for communication efficiency.
Since low-rank optimization is an NP-hard problem, we replace the rank function with the nuclear norm following \citet{zhang2025cut}:
\begin{equation}\small
    \operatorname*{arg\,min}_{\tilde{\mathcal{\boldsymbol{A}}}_\text{intra},\tilde{\mathcal{\boldsymbol{A}}}_\text{inter}}\;\sum_{t=1}^{T}{\|\tilde{\boldsymbol{A}}_\text{intra}^{(t)}\|_*} + \sum_{t=2}^{T}{\|\tilde{\boldsymbol{A}}_\text{inter}^{(t)}\|_*}
\end{equation}

Then in each intra- and inter-round adjacency matrix, we select the edges corresponding to the largest elements with a ratio of $1 - \alpha$ to keep and delete the rest of them:
\begin{equation}\label{eq:select_edges}\small
\begin{aligned}
    \mathcal{\boldsymbol{E}}_\Delta = &\bigcup_t \; \left(\boldsymbol{E}_\text{intra}^{(t)}\setminus\texttt{TopkEdges}\left(\tilde{\boldsymbol{A}}_\text{intra}^{(t)}, 1 - \beta\right)\right) \\
    \cup &\bigcup_t \; \left(\boldsymbol{E}_\text{inter}^{(t)}\setminus\texttt{TopkEdges}\left(\tilde{\boldsymbol{A}}_\text{inter}^{(t)}, 1 - \beta\right)\right)
\end{aligned}
\end{equation}
where $\texttt{TopkEdges}(\cdot)$ is a function returning a set of edges corresponding to top-$k$ elements in the adjacency matrix, and $\beta$ is the edge dropout rate.

As shown in the third row in Figure \ref{main_fig}, the final communication graph $\hat{\mathcal{\boldsymbol{G}}}$ is sampled from the optimized $\tilde{\mathcal{\boldsymbol{A}}}_\text{intra},\tilde{\mathcal{\boldsymbol{A}}}_\text{inter} $ using \texttt{DAGSample}.
The agents then engage in multi-round communication based on the topology defined in $\hat{\mathcal{\boldsymbol{G}}}$, and their individual responses are then aggregated to generate the final answer.
More detailed information on the optimization procedure is presented Appendix \ref{sec:algorithm}.

\subsection{Discussion}

Previous studies~\citep{you2020drawing,zhang2024two} have demonstrated that high-quality topological graphs can be trained using a limited number of samples. Therefore, in our approach, we employ only a small number of samples to update the parameters of the adjacency matrices for the communication graph, subsequently applying these matrices during two dropout processes. The number of these samples is merely in the order of tens, extracted from either the train or the validation set of the dataset, thereby maintaining the training cost and token consumption at relatively low levels.

\section{Experiments}

\begin{table*}[ht]
  \centering
  \scalebox{0.76}{
\begin{tabular}{l|*{8}{rr}}
\Xhline{1.2pt}
\rowcolor{CadetBlue!20} 

\multirow{2}{*}{\textbf{Method}} 
  & \multicolumn{2}{c}{\textbf{MMLU}} 
  & \multicolumn{2}{c}{\textbf{GSM8K}} 
  & \multicolumn{2}{c}{\textbf{AQuA}} 
  & \multicolumn{2}{c}{\textbf{MultiArith}} 
  & \multicolumn{2}{c}{\textbf{SVAMP}} 
  & \multicolumn{2}{c}{\textbf{HumanEval}} 
  & \multicolumn{2}{c}{\textbf{Avg.}}  \\
\cline{2-15}

 & \textbf{Ptok.} & \textbf{Ctok.} 
  & \textbf{Ptok.} & \textbf{Ctok.} 
  & \textbf{Ptok.} & \textbf{Ctok.} 
  & \textbf{Ptok.} & \textbf{Ctok.} 
  & \textbf{Ptok.} & \textbf{Ctok.} 
  & \textbf{Ptok.} & \textbf{Ctok.} 
  & \textbf{Ptok.} & \textbf{Ctok.}  \\
\Xhline{1.2pt}

Vanilla        & 99K & 44K & 142K & 209K & 32K & 89K & 16K & 15k & 27K & 19K & 91K & 25K & 68K & 67K \\
CoT            & 129K & 70K & 154K & 337K & 34K & 105K & 18K & 23k & 30K & 50K & 93K & 29K & 76K & 102K \\
$\text{MAS}_{round=1}$   & 1.4M & 355K & 8.5M & 1.9M & 1.1M & 390K & 1.1M & 218K & 1.9M & 402K & 581K & 134K & 2.4M & 567K \\
$\text{MAS}_{round=T}$   & 1.6M & 387K & 16M & 3.4M & 2.4M & 745K & 2.1M & 388K & 3.7M & 721K & 2.6M & 492K & 4.7M & 1.0M \\
AgentPrune     & 1.3M & 367K & 15M & 3.6M & 2.0M & 759K & 1.9M & 393K & 3.4M & 714K & 1.5M & 465K & 4.2M & 1.0M \\
AgentDropout   & 1.1M & 333K & 12M & 2.8M & 1.3M & 634K & 1.4M & 312K & 2.6M & 594K & 1.1M & 359K & 3.3M & 839K \\

\Xhline{1.2pt}
\end{tabular}
}
\caption{Token consumption comparison in the Llama model. \textbf{Ptok.} denotes the number of prompt tokens for the agents, while \textbf{Ctok.} represents the number of completion tokens generated by the agents.}
\label{main_tokens}
\end{table*}

\begin{table*}[ht]
  \centering
  \scalebox{0.8}{
\begin{tabular}{l|l|*{9}{c}}
\Xhline{1.2pt}
\rowcolor{CadetBlue!20} 
\textbf{Graph}
  & \textbf{Method}
  & \textbf{MMLU} 
  & \textbf{GSM8K} 
  & \textbf{AQuA} 
  & \textbf{MultiArith} 
  & \textbf{SVAMP} 
  & \textbf{HumanEval} 
  & \textbf{Avg.}  
  & \textbf{Ptok.}
  & \textbf{Ctok.}\\
\Xhline{1.2pt}

\multirow{3}{*}{Layered}
& MAS                & 57.52 & 71.72 & 47.92 & 87.26 & 78.84 & 52.89 & 66.13 & 4.3M & 1.1M \\
& AgentPrune         & 59.48 & 71.09 & 46.67 & 88.93 & 78.67 & 56.20 & 66.84 & 3.7M & 1.0M \\
& AgentDropout       & \textbf{61.44} & \textbf{72.27} & \textbf{49.17} & \textbf{90.25} & \textbf{79.75} & \textbf{57.02} & \textbf{68.32} & \textbf{2.8M} & \textbf{797K} \\

\Xhline{1.2pt}
\multirow{3}{*}{Random}
& MAS                & 62.75 & 70.63 & 46.67 & 87.22 & 77.17 & 50.41 & 65.81 & 4.2M & 1.0M \\
& AgentPrune         & 63.40 & 69.61 & 44.17 & 91.11 & 78.34 & 51.24 & 66.31 & 3.4M & 1.0M \\
& AgentDropout       & \textbf{64.71} & \textbf{72.03} & \textbf{47.50} & \textbf{91.67} & \textbf{79.50} & \textbf{53.72} & \textbf{68.19} & \textbf{2.7M} & \textbf{834K} \\

\Xhline{1.2pt}

\end{tabular}
}
\caption{Performance and average token consumption achieved with different initial communication graph topological structures.}
\label{graph_stability}
\end{table*}

\subsection{Experimental Setup}
\label{setup}
\paragraph{Models and Benchmarks}
We conduct experiments on MAS based on models of varying sizes, namely Meta-Llama3-8B-Instruct~\citep{llama3modelcard}, Qwen2.5-72B-Instruct~\citep{qwen2.5}, and Deepseek-V3-671B-Instruct~\citep{liu2024deepseek}. We evaluate the general reasoning capabilities of our method using the MMLU~\citep{hendrycks2021measuring} dataset; the mathematical reasoning abilities using the GSM8k~\citep{cobbe2021gsm8k}, MultiArith~\citep{roy-roth-2015-solving}, AQuA~\citep{patel-etal-2021-nlp}, and SVAMP~\citep{ling2017program} datasets; and the code generation capabilities using the HumanEval~\citep{chen2021codex} dataset.

\paragraph{Baselines}
For tasks performed by a single agent, we compare with the direct reasoning method, known as Vanilla, and the Chain-of-Thought~(\citealp[CoT,][]{wei2022chain}) approach. In the context of multi-agent collaboration, for methods that only communicate within intra-round, we compare the single-round dialogue MAS, denoted as $\text{MAS}_{\text{round}=1}$. For methods that communicate both within intra-round and across inter-round, we compare $\text{MAS}_{\text{round}=T}$ and AgentPrune~\citep{zhang2025cut}, where $\mathrm{round}$ represents the number of inter-round communication rounds among agents. In particular, both AgentPrune and AgentDropout are optimized based on $\text{MAS}_{\text{round}=T}$, allowing for a more intuitive comparison of these methods.

\paragraph{Implementation Details}
For experiments with Llama3 and Qwen2.5, we perform inference using vllm on Nvidia A800 GPUs. For Deepseek-V3, we utilize the Deepseek API to evaluate each method. We set the temperature to 1, the sample times $M$ in \{10, 20\}, the learning rate $\eta$ to 0.1, and the note and edge dropout rate $\alpha$ and $\beta$ in \{0.1,0.2\}, respectively. For all benchmarks, the number of training samples we extracted is 40. For general reasoning tasks and mathematical tasks, we set the number of communication rounds $T=2$, while for code generation tasks, we configure $T=4$. To ensure a fair comparison, we directly adopt the agent configuration files from AgentPrune.

\begin{table*}[ht]
  \centering
  \scalebox{0.8}{
\begin{tabular}{l|*{9}{c}}
\Xhline{1.2pt}
\rowcolor{CadetBlue!20} 
\textbf{Dropout Rate}
  & \textbf{MMLU} 
  & \textbf{GSM8K} 
  & \textbf{AQuA} 
  & \textbf{MultiArith} 
  & \textbf{SVAMP} 
  & \textbf{HumanEval} 
  & \textbf{Avg.}    
  & \textbf{Ptok.}  
  & \textbf{Ctok.}  \\
\Xhline{1.2pt}

\multirow{1}{*}{0.8}  & 57.52 & 70.86 & 46.67 & 89.44 & 78.67 & 52.89 & 66.01 & \textbf{856K} & \textbf{230K} \\
\multirow{1}{*}{0.6}  & 60.78 & 71.09 & 45.42 & 89.44 & \textbf{80.37} & 50.41 & 66.25 & 1.3M & 434K \\
\multirow{1}{*}{0.4}  & 60.13 & \textbf{73.36} & 44.99 & 90.00 & 78.33 & 52.07 & 66.48 & 1.9M & 648K \\
\multirow{1}{*}{0.2}  & \textbf{62.75} & 73.13 & \textbf{47.78} & \textbf{92.59} & 80.11 & \textbf{55.84} & \textbf{68.70} & 3.3M & 839K \\

\Xhline{1.2pt}
\end{tabular}
}
\caption{Impact of varying dropout rates on the performance and average token consumption of AgentDropout.}
\label{influence_dropout}
\end{table*}

\begin{table*}[ht]
  \centering
  \scalebox{0.8}{
\begin{tabular}{l|cc|*{7}{c}}
\Xhline{1.2pt}
\rowcolor{CadetBlue!20} 
\textbf{Method}
  & \textbf{Edge DR.} 
  & \textbf{Node DR.}
  & \textbf{MMLU} 
  & \textbf{GSM8K} 
  & \textbf{AQuA} 
  & \textbf{MultiArith} 
  & \textbf{SVAMP} 
  & \textbf{HumanEval} 
  & \textbf{Avg.}  \\
\Xhline{1.2pt}

MAS  & \textcolor{darksalmon}{\XSolidBrush} & \textcolor{darksalmon}{\XSolidBrush} & 60.13 & 71.48 & 45.41 & 90.56 & 77.56 & 49.17 & 65.72 \\
\hline
\multirow{2}{*}{Random Dropout}  & \textcolor{darksalmon}{\XSolidBrush} & \textcolor{green(pigment)}{\Checkmark} & 61.72 & 70.53 & 43.75 & 92.23 & 79.00 & 52.89 & 66.69 \\
& \textcolor{green(pigment)}{\Checkmark} & \textcolor{darksalmon}{\XSolidBrush} & 60.13 & 71.41 & 46.05 & 90.27 & 77.67 & 51.23 & 66.13 \\
\hline
\multirow{2}{*}{Learned Dropout}
& \textcolor{darksalmon}{\XSolidBrush} & \textcolor{green(pigment)}{\Checkmark} & \textbf{63.76} & 70.70 & 46.25 & 90.28 & 79.17 & 50.83 & 66.83 \\
& \textcolor{green(pigment)}{\Checkmark} & \textcolor{darksalmon}{\XSolidBrush} & 61.87 & 70.63 & 47.08 & 90.00 & 78.03 & 48.76 & 66.06 \\

\hline
Single Learning  & \textcolor{green(pigment)}{\Checkmark} & \textcolor{green(pigment)}{\Checkmark} & 60.78 & 72.58 & 46.25 & 91.67 & 78.83 & 48.73 & 66.47 \\
AgentDropout & \textcolor{green(pigment)}{\Checkmark} & \textcolor{green(pigment)}{\Checkmark} & 62.75 & \textbf{73.13} & \textbf{47.78} & \textbf{92.59} & \textbf{80.11} & \textbf{55.84} & \textbf{68.70} \\

\Xhline{1.2pt}
\end{tabular}
}
\caption{Performance achieved with different dropout strategies. For Random Dropout, \textcolor{green(pigment)}{\Checkmark} denotes the random dropout strategy is applied to the marked dropout step, while the dropout strategy in the other step remains unchanged, as per AgentDropout.}
\label{dropout_strategy}
\end{table*}

\begin{table}[t]
  \centering
  \scalebox{0.7}{
\begin{tabular}{l|*5{c}}
\Xhline{1.2pt}
\rowcolor{CadetBlue!20} 
\textbf{Method} 
  & \textbf{GSM8K} 
  & \textbf{AQuA} 
  & \textbf{MultiArith} 
  & \textbf{SVAMP} 
  & \textbf{Avg.}\\
\Xhline{1.2pt}

\multirow{1}{*}{GSM8K}      & 73.13 & 46.67 & 92.22 & \textbf{80.33} & 73.09 \\
\multirow{1}{*}{AQuA}       & \textbf{73.84} & \textbf{47.78} & \textbf{93.13} & 78.21 & 73.24 \\
\multirow{1}{*}{MultiArith} & 71.41 & 47.50 & 92.59 & 78.33 & 72.46 \\
\multirow{1}{*}{SVAMP}      & 70.86 & 47.50 & 91.95 & 80.11 & 72.61 \\
\hdashline
AgentPrune         & 71.02 & 47.22 & 90.00 & 78.34 & 71.65 \\
AgentDropout       & 73.13 & \textbf{47.78} & 92.59 & 80.11 & \textbf{73.40} \\

\Xhline{1.2pt}
\end{tabular}
}
\caption{Performance comparison using different combinations of training and test sets. The row headers indicate the training set used for graph topology learning, and the column headers represent the test set. }
\label{domain_robustness}
\end{table}

\subsection{Main Result}
\paragraph{AgentDropout outperforms other baselines on multiple benchmarks.}
As shown in Table~\ref{main_result}, our method consistently outperforms both the single-agent method CoT and the state-of-the-art multi-agent communication method AgentPrune across all benchmarks, regardless of the model scale. Specifically, using the Llama3 model, AgentDropout improves accuracy by 2.19 compared to the SOTA method. In experiments with Qwen2.5 and Deepseek-V3, the accuracy improves by 0.77 and 0.47, respectively. These results demonstrate the effectiveness of AgentDropout. A case study of the reasoning results is provided in Appendix~\ref{sec:case}.

\paragraph{AgentDropout improves the stability of MAS.}
In Table~\ref{main_result}, it can be observed that when the model capability is relatively weak (e.g., Llama3), optimization methods for single agents, such as CoT, exhibit more significant performance improvements, whereas the performance gains from multi-agent interaction and collaboration are less stable, as seen in MAS and AgentPrune. 
However, when the model capability is sufficiently strong, the performance improvements brought by single-agent optimization methods diminish, while multi-agent communication methods achieve more stable enhancements. Notably, for models of different scales, AgentDropout consistently enables stable performance improvements in MAS, demonstrating its model robustness.

\paragraph{AgentDropout reduces the token cost of multi-agent communication and cooperation.}
As shown in Table~\ref{main_tokens}, compared to the multi-round communication method $\text{MAS}_{\text{round}=T}$, AgentDropout significantly reduces token consumption. In contrast to AgentPrune, which primarily reduces prompt token consumption, AgentDropout not only further decreases prompt tokens but also significantly reduces completion tokens, revealing its token efficiency. For more details, please refer to Appendix~\ref{apd:tokenefficiency}.

\subsection{Analysis}
In this part, we conducted experiments using the Llama3 model, while maintaining the same experimental settings as described in Section~\ref{setup}.
\paragraph{Structure Robustness}
To verify the stability of AgentDropout when initialized with different topological structures, we also conducted supplementary experiments on communication graphs initialized with layered and random graphs, which are introduced in Appendix~\ref{sec:graph}. 
As shown in Table~\ref{graph_stability}, AgentDropout consistently enhances the performance and reduces the average consumption of prompt tokens and completion tokens of MAS across various graph structures, which further demonstrates its robustness.

\paragraph{Effect of Dropout Rate}
To investigate the impact of different dropout rates on the performance of MAS, we conducted experiments with varying values of $\alpha$ and $\beta$, as shown in Table~\ref{influence_dropout}. 
The results demonstrate that higher dropout rates lead to a corresponding decline in the average performance of MAS. 
However, compared with the SOTA methods, AgentDropout still achieves superior average performance across different dropout rates, which substantiates the effectiveness of dropout. In extreme cases, higher dropout rates can be set to ensure efficiency. 
Nevertheless, for tasks such as AQuA and HumanEval, MAS exhibits lower sensitivity to dropout rates, suggesting that future work could focus on designing more appropriate MAS frameworks for such scenarios.

\paragraph{Different Dropout Strategies}
To verify the effectiveness of our dropout strategy, we conducted a series of comparative experiments.
First, we performed an ablation study on the overall AgentDropout methodology, comparing it against scenarios that use only Node Dropout and only Edge Dropout.
Additionally, to rule out the possibility that the performance improvement is simply due to the shortening of the reasoning chain, we conduct an experiment where the proposed dropout mechanism is replaced with a random dropout approach in either the Node Dropout or Edge Dropout stage while keeping all other processes unchanged.

As shown in the ``Learned Dropout'' block of Table~\ref{dropout_strategy}, applying either Node Dropout or Edge Dropout alone results in improvements compared to the vanilla MAS, indicating the reasonableness of the dropout approaches.
However, the simultaneous application of both Node Dropout and Edge Dropout, referred to as ``AgentDropout'', yields the optimal performance.
In the random dropout experiments, randomly dropping either edges or nodes within the AgentDropout framework leads to performance degradation, further substantiating the effectiveness of our proposed dropout strategy.

\paragraph{Training Strategy}
To validate the effectiveness of our two-round adjacency matrix learning strategy (one for Node Dropout and one for Edge Dropout), we conducted comparative experiments with single-round learning. Specifically, we trained both intra- and inter-round adjacency matrices simultaneously, and then directly applied them to perform Node and Edge Dropout on MAS. The experimental results are presented in Table~\ref{dropout_strategy}. The performance of AgentDropout significantly outperforms that of single-round learning. This improvement can be attributed to the fact that after Node Dropout, the communication graph generates new inference chains, necessitating an additional training round to effectively implement further dropout.

\paragraph{Domain Transferability}
Our approach requires some training instances of the current task to conduct communication graph optimization and Node/Edge dropout.
However, in the evaluation phase, there may be no training data available that directly corresponds to the current test data.
To this point, we conduct a domain robustness test, assessing how well our system works when only roughly similar training data is available.
Specifically, we utilize dataset A to optimize the communication topology graph and evaluate the trained MAS on dataset B. We conduct experiments on mathematics-related datasets.

As shown in Table~\ref{domain_robustness}. When using other datasets as training samples, AgentDropout still maintains robust performance. When trained on the more challenging AQuA dataset, it achieves an average performance improvement of 1.59 over the SOTA method, while for the simpler MultiArith dataset, the average improvement is 0.81. This demonstrates that the communication subgraphs learned by AgentDropout for tasks of the same category exhibit strong generalizability. Moreover, the effectiveness of AgentDropout is positively correlated with the difficulty of the training samples. Specifically, for tasks of the same type, the more challenging the task, the stronger the generalizability learned by MAS. Therefore, when applying AgentDropout to few-shot tasks, selecting similar datasets for training is recommended, which further validates its domain robustness.

\section{Related Work}
\subsection{LLM-based MAS}
MAS based on LLMs has emerged as a crucial tool for tackling complex tasks across various domains. Early research has demonstrated the significant potential of LLM-driven agents in collaboratively solving problems through multi-agent debates~\citep{liang-etal-2024-encouraging,du2024improving}, laying the foundation for several research directions, including role-playing in complex reasoning~\citep{wang-etal-2024-unleashing,chen2024agentverse}, collaborative software development~\citep{ishibashi-and-nishimura-2024self,qian-etal-2024-chatdev,hong2024metagpt}, and embodied agent interactions~\citep{10610855,zhang2024building,guo2024embodied}. Recent studies have shown that increasing the number and diversity of agents can enhance the overall performance of MAS~\citep{wang2025mixtureofagents,li2024more,chen2025internet}. As LLM-based MAS continues to scale in size and complexity, the issues of computational cost and communication efficiency have become more pronounced~\citep{chen-etal-2024-beyond-natural,li-etal-2024-improving-multi}. Recent work has proposed methods to optimize MAS efficiency, such as Optima~\citep{chen2024optima} and AgentPrune~\citep{zhang2025cut}. In this paper, we address the problem of communication redundancy by applying dropout to nodes and edges in the graph structure of MAS, significantly improving computational cost and communication efficiency in practical applications.

\subsection{MAS as Graphs}
In recent years, researchers have proposed learning graph connections to enhance the collaborative performance of MAS~\citep{hu2024learning,pesce2023learning}. Prior to the widespread adoption of LLMs, existing methods explored optimal communication graph structures for MAS using techniques such as graph diffusion, weighted GNNs, and transformers~\citep{liu2022temporal,pesce2023learning,hu2024learning}. For LLM-based MAS, previous work has also attempted to use graphs to represent interactions between agents. 
For instance, methods like ChatEval~\citep{chan2024chateval} and AutoGen~\citep{wu2023autogen} implicitly incorporate graph structures to model simultaneous conversations, while other approaches such as STOP~\citep{zelikman2023self} and DSPy~\citep{khattab2024dspy} optimize prompt and reasoning structures. Additionally, methods like MacNet~\citep{qian2025scaling}, GPTSwarm~\citep{zhuge2024gptswarm}, EvoMAC~\citep{hu2025selfevolving} and DyLAN~\citep{liu2024a} optimize the topology of agents communication.

\section{Conclusion}
In this paper, inspired by management theories that roles in efficient teams are often dynamically adjusted, we propose AgentDropout, a novel topology optimization method for MAS with domain transferability and structure robustness. AgentDropout dynamically adjusts the participating agents and communication links among agents in each round, allowing for more flexible and adaptive team configurations. Experiments on a series of tasks demonstrate that AgentDropout not only reduces token consumption but also consistently enhances the performance of MAS across various tasks. This provides valuable insights for future exploration of larger-scale multi-agent interactions.

\section*{Limitations}
The tasks validated in this paper are relatively limited and focus mainly on conventional reasoning tasks, mathematical tasks, and code generation tasks. To better evaluate the generalizability of AgentDropout, additional experiments across a broader and more diverse range of tasks are essential. Furthermore, although AgentDropout demonstrates promising efficiency and performance, the current framework remains constrained by predefined agent roles and prompts. When addressing tasks in unfamiliar domains, the careful design of new initial roles and prompts is still crucial. Future work could also explore integrating methods for automatically designing roles and prompts to enhance flexibility. Additionally, although AgentDropout reduces the token consumption of MAS, it remains relatively high, indicating significant opportunities for further optimization.

\bibliography{custom}

\begin{algorithm*}[!ht]
\DontPrintSemicolon
\SetAlgoLined
\LinesNumbered
\SetKwFunction{FPruneNode}{PruneNode}
\SetKwFunction{FPruneEdge}{PruneEdge}
\SetKwFunction{FLoss}{StrengthLoss}
\SetKwFunction{FEdgeLoss}{EdgeLoss}
\SetKwFunction{CalStrength}{CalStrength}
\SetKwFunction{DAGSample}{DAGSample}
\SetKwFunction{Optimize}{Update}
\SetKwFunction{NodeDropout}{NodeDropout}
\SetKwFunction{EdgeDropout}{EdgeDropout}
\SetKwFunction{TopkEdges}{TopkEdges}
\SetKwFunction{SelectNodes}{SelectNodes}
\SetKwFunction{SelectEdges}{SelectEdges}
\KwIn{Initial communication graph $\mathcal{\boldsymbol{G}}=\left(\bigcup_{t} \boldsymbol{V}^{(t)}, \left(\bigcup_{t} \boldsymbol{E}_\text{intra}^{(t)}\right) \cup \left(\bigcup_{t} \boldsymbol{E}_\text{inter}^{(t)}\right), \bigcup_{t} \boldsymbol{F}^{(t)}\right)$, Initial adjacency matrices $\tilde{\mathcal{\boldsymbol{A}}}_\text{intra}=\left\{\tilde{\mathcal{\boldsymbol{A}}}_\text{intra}^{(t)}\right\}_t, \tilde{\mathcal{\boldsymbol{A}}}_\text{inter}=\left\{\tilde{\mathcal{\boldsymbol{A}}}_\text{inter}^{(t)}\right\}_t$, Training steps $K$, Sampling times $M$, Learning rate $\eta$, Node dropout rate $\alpha$, Edge dropout rate $\beta$}
\KwOut{Communication graph after dropout $\hat{\mathcal{\boldsymbol{G}}}$}

\tcp{Node Dropout}

$\texttt{Initialize}\left(\tilde{\mathcal{\boldsymbol{A}}}_\text{intra},\tilde{\mathcal{\boldsymbol{A}}}_\text{inter}\right)$ \tcp{Set the elements to 0.5}

\For{\rm{iter} \textit{k} $\leftarrow$ 1 \KwTo \textit{K}}{
$\left\{\mathcal{\boldsymbol{G}}_m\right\}_{m=1}^M \gets \left\{\texttt{DAGSample}\left(\tilde{\mathcal{\boldsymbol{A}}}_\text{intra},\tilde{\mathcal{\boldsymbol{A}}}_\text{inter}\right)\right\}_{m=1}^M$ \tcp{Sample $M$ communication graphs}
$\tilde{\mathcal{\boldsymbol{A}}}_\text{intra}\gets \tilde{\mathcal{\boldsymbol{A}}}_\text{intra} + \eta \cdot \frac{1}{M}\sum_{m=1}^M\mu\left(\mathcal{\boldsymbol{G}}_m\right) \nabla_{\tilde{\mathcal{\boldsymbol{A}}}_\text{intra}}\log\left(p_{\tilde{\mathcal{\boldsymbol{A}}}_\text{intra}}\left(\mathcal{\boldsymbol{G}}_m\right)\right)$ \tcp{Optimize $\tilde{\mathcal{\boldsymbol{A}}}_\text{intra}$ according to Equation (\ref{eq:node_dropout_optimization})}
}

$\mathcal{\boldsymbol{V}}_\Delta = \bigcup_{t} \; \left(\boldsymbol{V}^{(t)}\setminus\texttt{TopkNodes}\left(\tilde{\boldsymbol{A}}_\text{inter}^{(t)}, 1-\alpha\right)\right)$ \tcp{Select nodes with small degrees}
$\tilde{\mathcal{\boldsymbol{A}}}_\text{intra},\tilde{\mathcal{\boldsymbol{A}}}_\text{inter} \gets \NodeDropout\left(\tilde{\mathcal{\boldsymbol{A}}}_\text{intra},\tilde{\mathcal{\boldsymbol{A}}}_\text{inter} ,\mathcal{\boldsymbol{V}}_\Delta\right)$ \tcp{Remove nodes from the graph}

\tcp{Edge Dropout}

$\texttt{Initialize}\left(\tilde{\mathcal{\boldsymbol{A}}}_\text{intra},\tilde{\mathcal{\boldsymbol{A}}}_\text{inter}\right)$

\For{\rm{iter} \textit{k} $\leftarrow$ 1 \KwTo \textit{K}}{
$\left\{\mathcal{\boldsymbol{G}}_m\right\}_{m=1}^M \gets \left\{\texttt{DAGSample}\left(\tilde{\mathcal{\boldsymbol{A}}}_\text{intra},\tilde{\mathcal{\boldsymbol{A}}}_\text{inter}\right)\right\}_{m=1}^M$ \tcp{Sample $M$ communication graphs}

$\left\{\tilde{\boldsymbol{A}}_\text{intra},\tilde{\boldsymbol{A}}_\text{inter}\right\}\gets \left\{\tilde{\boldsymbol{A}}_\text{intra},\tilde{\boldsymbol{A}}_\text{inter}\right\} + \eta \cdot \frac{1}{M}\sum_{m=1}^M\mu\left(\mathcal{\boldsymbol{G}}_m\right) \nabla_{\tilde{\mathcal{\boldsymbol{A}}}_\text{intra},\tilde{\mathcal{\boldsymbol{A}}}_\text{inter}}\log\left(p_{\tilde{\mathcal{\boldsymbol{A}}}_\text{intra},\tilde{\mathcal{\boldsymbol{A}}}_\text{inter}}\left(\mathcal{\boldsymbol{G}}_m\right)\right)$ \tcp{Optimize $\tilde{\mathcal{\boldsymbol{A}}}$ according to Equation (\ref{eq:edge_dropout_optimization})}
}

$\mathcal{\boldsymbol{E}}_\Delta = \bigcup_t \; \left(\boldsymbol{E}_\text{intra}^{(t)}\setminus\texttt{TopkEdges}\left(\tilde{\boldsymbol{A}}_\text{intra}^{(t)}, 1 - \beta\right)\right)
\cup \bigcup_t \; \left(\boldsymbol{E}_\text{inter}^{(t)}\setminus\texttt{TopkEdges}\left(\tilde{\boldsymbol{A}}_\text{inter}^{(t)}, 1 - \beta\right)\right)$ \tcp{Select edges with small weights}
$\tilde{\mathcal{\boldsymbol{A}}}_\text{intra},\tilde{\mathcal{\boldsymbol{A}}}_\text{inter} \gets\EdgeDropout\left(\tilde{\mathcal{\boldsymbol{A}}}_\text{intra},\tilde{\mathcal{\boldsymbol{A}}}_\text{inter} ,\mathcal{\boldsymbol{E}}_\Delta\right)$ \tcp{Remove edges from the graph}
$\hat{\mathcal{\boldsymbol{G}}}\gets\DAGSample\left(\tilde{\mathcal{\boldsymbol{A}}}_\text{intra},\tilde{\mathcal{\boldsymbol{A}}}_\text{inter} \right)$ \tcp{Sample the final communication DAG}
\Return $\hat{\mathcal{\boldsymbol{G}}}$
\caption{Communication Graph Dropout}
\label{algo:matrix_opt}
\end{algorithm*}

\appendix

\begin{table*}[ht]
  \centering
  \scalebox{0.76}{
\begin{tabular}{l|*{8}{rr}}
\Xhline{1.2pt}
\rowcolor{CadetBlue!20} 

\multirow{2}{*}{\textbf{Method}} 
  & \multicolumn{2}{c}{\textbf{MMLU}} 
  & \multicolumn{2}{c}{\textbf{GSM8K}} 
  & \multicolumn{2}{c}{\textbf{AQuA}} 
  & \multicolumn{2}{c}{\textbf{MultiArith}} 
  & \multicolumn{2}{c}{\textbf{SVAMP}} 
  & \multicolumn{2}{c}{\textbf{HumanEval}} 
  & \multicolumn{2}{c}{\textbf{Avg.}}  \\
\cline{2-15}

 & \textbf{Ptok.} & \textbf{Ctok.} 
  & \textbf{Ptok.} & \textbf{Ctok.} 
  & \textbf{Ptok.} & \textbf{Ctok.} 
  & \textbf{Ptok.} & \textbf{Ctok.} 
  & \textbf{Ptok.} & \textbf{Ctok.} 
  & \textbf{Ptok.} & \textbf{Ctok.} 
  & \textbf{Ptok.} & \textbf{Ctok.}  \\
\Xhline{1.2pt}

\multicolumn{15}{c}{Base model: Llama3-8B-Instruct} \\
\hline
Vanilla        & 99K & 44K & 142K & 209K & 32K & 89K & 16K & 15k & 27K & 19K & 91K & 25K & 68K & 67K \\
CoT            & 129K & 70K & 154K & 337K & 34K & 105K & 18K & 23k & 30K & 50K & 93K & 29K & 76K & 102K \\
$\text{MAS}_{round=1}$   & 1.4M & 355K & 8.5M & 1.9M & 1.1M & 390K & 1.1M & 218K & 1.9M & 402K & 581K & 134K & 2.4M & 567K \\
$\text{MAS}_{round=T}$   & 1.6M & 387K & 16M & 3.4M & 2.4M & 745K & 2.1M & 388K & 3.7M & 721K & 2.6M & 492K & 4.7M & 1.0M \\
AgentPrune     & 1.3M & 367K & 15M & 3.6M & 2.0M & 759K & 1.9M & 393K & 3.4M & 714K & 1.5M & 465K & 4.2M & 1.0M \\
AgentDropout   & 1.1M & 333K & 12M & 2.8M & 1.3M & 634K & 1.4M & 312K & 2.6M & 594K & 1.1M & 359K & 3.3M & 839K \\
\hline

\multicolumn{15}{c}{Base model: Qwen2.5-72B-Instruct} \\
\hline
Vanilla        & 99K & 42K & 142K & 232K & 32K & 86K & 16K & 14k & 27K & 18K & 91K & 24K & 68K & 69K \\
CoT            & 129K & 71K & 154K & 340K & 34K & 113K & 18K & 28k & 30K & 56K & 93K & 26K & 76K & 106K \\
$\text{MAS}_{round=1}$   & 966K & 315K & 9.0M & 1.8M & 1.4M & 503K & 1.1M & 168K & 1.8M & 293K & 271K & 42K & 2.4M & 520K \\
$\text{MAS}_{round=T}$   & 2.2M & 663K & 18M & 3.5M & 2.9M & 988K & 2.0M & 314K & 3.4M & 546K & 814K & 121K & 4.9M & 1.0M \\
AgentPrune     & 2.0M & 665K & 15M & 3.3M & 2.4M & 995K & 1.8M & 313K & 3.1M & 517K & 512K & 83K & 4.1M & 979K \\
AgentDropout   & 1.3M & 456K & 11M & 2.6M & 2.0M & 779K & 1.6M & 234K & 2.4M & 466K & 421K & 77K & 3.1M & 769K \\
\hline

\multicolumn{15}{c}{Base model: Deepseek-V3-671B-Instruct} \\
\hline
Vanilla        & 99K & 39K & 137K & 233K & 30K & 79K & 16K & 14K & 26K & 28K & 92K & 27K & 67K & 70K \\
CoT            & 129K & 65K & 148K & 289K & 34K & 98K & 17K & 23K & 29K & 50K & 93K & 27K & 75K & 92K \\
$\text{MAS}_{round=1}$   & 845K & 247K & 7.9M & 1.6M & 1.2M & 444K & 951K & 157K & 1.6M & 258K & 339K & 71K & 2.1M & 463K \\
$\text{MAS}_{round=T}$   & 2.1M & 575K & 16M & 3.2M & 2.7M & 915K & 1.9M & 285K & 3.0M & 456K & 1.3M & 286K & 4.5M & 953K \\
AgentPrune     & 1.6M & 582K & 13M & 3.2M & 2.2M & 902K & 1.7M & 296K & 2.9M & 498K & 649K & 206K & 3.7M & 947K \\
AgentDropout   & 1.0M & 328K & 11M & 2.7M & 1.6M & 847K & 1.4M & 237K & 2.3M & 399K & 469K & 169K & 3.0M & 780K \\
\Xhline{1.2pt}
\end{tabular}
}
\caption{Token consumption comparison between AgentDropout and other baseline reasoning techniques. \textbf{Ptok.} denotes the number of prompt tokens for the agents, while \textbf{Ctok.} represents the number of completion tokens generated by the agents.}
\label{apd_tokens}
\end{table*}

\section{Appendix}
\subsection{Optimization Algorithm of AgentDropout}
\label{sec:algorithm}
Algorithm \ref{algo:matrix_opt} presents the detailed pseudo code of AgentDropout.
The input parameters for this algorithm mainly include communication graph $\mathcal{\boldsymbol{G}}$ to be optimized, along with its intra- and inter-round adjacency matrices $\tilde{\mathcal{\boldsymbol{A}}}_\text{intra}, \tilde{\mathcal{\boldsymbol{A}}}_\text{inter}$.
The output is the optimized communication graph $\hat{\mathcal{\boldsymbol{G}}}$.

During NodeDropout, we first call the $\texttt{Initialize}(\cdot)$ function to initialize the adjacency matrices by setting the relevant elements to 0.5 (Line 1).
We then train the intra-round matrices to optimize the task performance of the communication graphs sampled from these matrices (Lines 2 \textasciitilde 5).
After the training process, we identify the nodes in each communication round with the smallest in- and out-degree (Line 6) and remove them, along with all the edges connected to them, by invoking the $\texttt{NodeDropout}(\cdot)$ function (Line 7).

During EdgeDropout, we initialize the intra- and inter-round adjacency matrices again (Line 8) and train both of them to optimize task performance and graph sparsity simultaneously (Lines 9 \textasciitilde 12).
Next, we identify the intra- and inter-round edges with the smallest weights in their respective matrices (Line 13) and remove them from the graph by calling the $\texttt{EdgeDropout}(\cdot)$ function (Line 14).
Finally, we sample the communication graph used in actual evaluation by calling the $\texttt{DAGSample}(\cdot)$ function, which samples the graph by treating the weights in the matrices as sampling possibility and guarantees the output to be a DAG graph (Line 15).

\subsection{Case Study}
\label{sec:case}
Given a multiple-choice physics question from the MMLU dataset, in the first round of interaction, the Knowledgeable Expert provides an initial analysis, emphasizing the need to search for information related to the equation and relevant atomic properties. The Critic critiques the response of the Knowledgeable Expert, pointing out the lack of practical analysis and application, and suggests further refinement of the reasoning. Upon receiving feedback, the Mathematician conducts a detailed analysis of the problem, noting that the equation cannot be applied to H$^-$, and discusses variations across different atoms, providing more specific reasoning. The Psychologist offers feedback from a psychological perspective, affirming the process of critique and revision, and encourages more comprehensive critical thinking. In the second round of interaction, the Historian briefly summarizes the reasons. The Critic revisits the perspectives of other agents, suggesting that the approach of Mathematicians to different atoms could be further improved, and emphasizes the impact of atomic properties on the equation. After receiving feedback, the Mathematician revises the analysis, clarifies which atoms can be addressed by modifying the application of the Rydberg equation, and concludes that the Hydride ion (H$^-$) is the only unsuitable option. The Historian in the first round and the Knowledgeable Expert in the second round are dropped out.

\subsection{Token Efficiency}
\label{apd:tokenefficiency}
As shown in Table~\ref{apd_tokens}, AgentDropout significantly reduces token consumption compared to the multi-round method $\text{MAS}_{\text{round}=T}$. Unlike AgentPrune, which reduces only prompt tokens, AgentDropout reduces both prompt and completion tokens. In the Llama3 model, it achieves a 21.4\% reduction in prompt tokens and 16.1\% in completion tokens. For Qwen2.5 and Deepseek-V3, the reductions are 24.4\% and 18.9\% in prompt tokens, and 21.4\% and 17.6\% in completion tokens, respectively, demonstrating its token efficiency.

\subsection{Different Communication Topologies}
\label{sec:graph}
\begin{figure*}[t]
\centering 
\includegraphics[height=5.2cm]{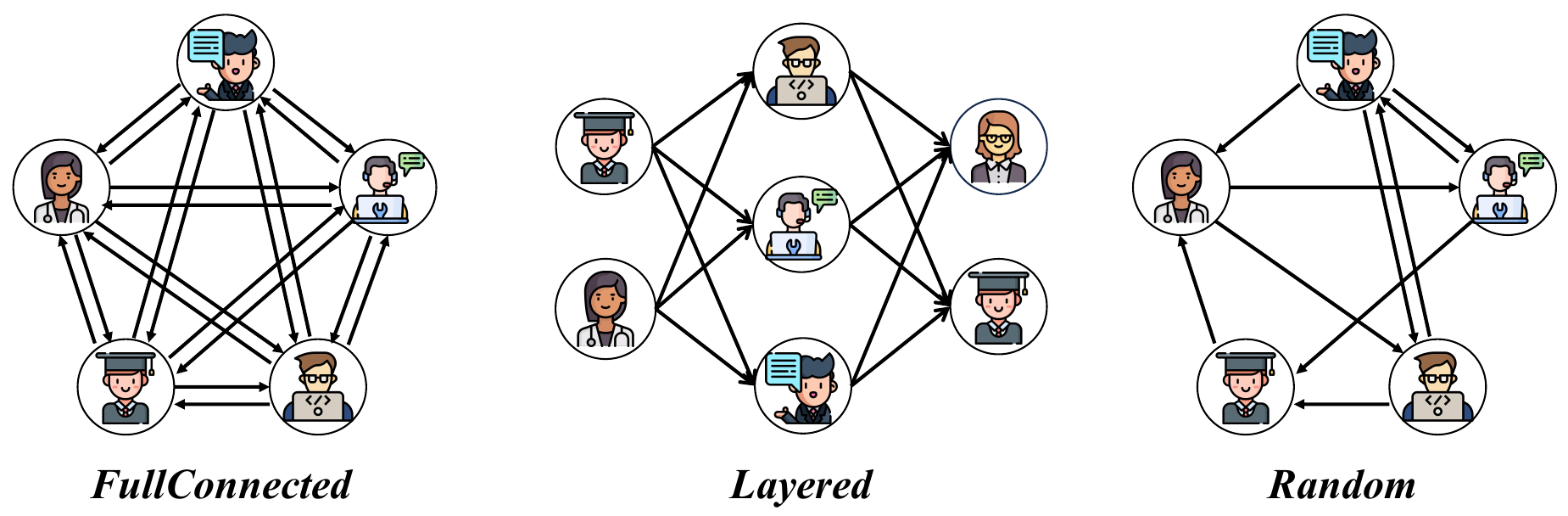}
\caption{Full connected, layer and random graph as demonstrations of intra-round communication topologies.}
\label{diff_graph}
\end{figure*}

In the main experiments, we initialize the intra-round communication graph using a fully connected graph. In the analysis experiments, we supplement this by conducting additional experiments with intra-round graphs initialized using layered graphs and random graphs, as shown in Figure~\ref{diff_graph}. For both fully connected graphs and random graphs, we first process the communication topology into a DAG using \texttt{DAGSample}, and then perform inference and communication based on the topological order~\citep{zhuge2024gptswarm,qian2025scaling}. For layered graphs, similar to multi-layer perceptrons, the agents in the first layer perform inference first, followed by message passing to the agents in the second layer, and so on.

\begin{table*}[t]
\begin{tcolorbox}[notitle, colback=white, colframe=softblue,
       boxrule=3pt, boxsep=0.5pt, enhanced, 
       shadow={3pt}{-3pt}{0pt}{opacity=1,mygrey},
       title={\large Task (1/12) \& \includegraphics[height=22pt]{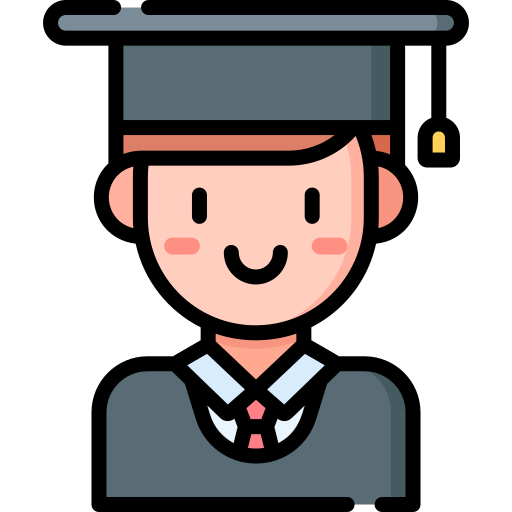} {\large Knowledgeable Expert 3FeF: Round 1 (2/12)}}]
\texttt{The Rydberg equation v = $R_H(1/n_1^2 - 1/n_2^2)$ accurately predicts the UV-visible emission spectrum of the hydrogen atom. A form of the Rydberg equation may also be used to predict the UV-visible emission for all of the following EXCEPT\\Option A: hydride ion, H-\\Option B: deuterium atom, D\\Option C: tritium atom, T\\Option D: helium cation, He+\\\rule[0mm]{\linewidth}{0.5pt} A fascinating topic!\\ \\To solve this problem, we need to search for entities related to the Rydberg equation and atomic properties. Here are the key entities to search in Wikipedia:\\ \\@Rydberg equation@\\@Hydrogen atom@\\@Hydride ion (H-)@\\@Deuterium atom (D)@\\@Tritium atom (T)@\\@Helium cation (He+)@\\These entities will provide us with the necessary information to determine which option(s) is/are incorrect.
}
\end{tcolorbox}
\end{table*}

\begin{table*}[t]
\begin{tcolorbox}[notitle, colback=white, colframe=softgreen,
       boxrule=3pt, boxsep=0.5pt, enhanced, 
       shadow={3pt}{-3pt}{0pt}{opacity=1,mygrey},
       title={\includegraphics[height=22pt]{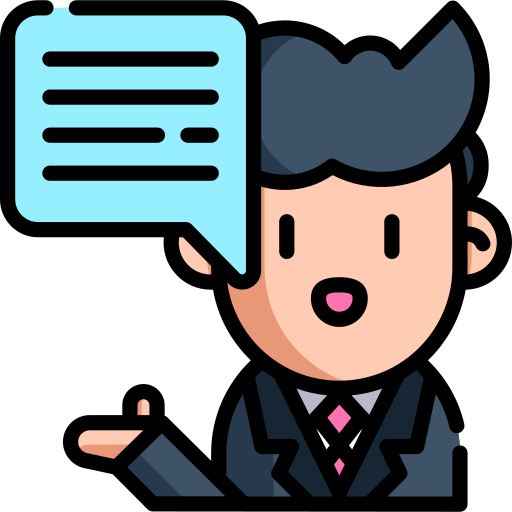} \large Critic 6PWE: Round 1 (3/12)}]
\texttt{As a critic, I'll analyze the output of Agent 3FeF and identify potential issues:\\\\1. Missing analysis: The output doesn't provide any actual analysis or reasoning for why the Rydberg equation might not apply to the mentioned options. It merely states that searching for entities related to the Rydberg equation and atomic properties is necessary, without actually applying those concepts to the problem.\\2. Lack of specific mention: The output doesn't mention a specific aspect of the Rydberg equation that might not be applicable to the mentioned options. It's unclear what part of the equation might be affected by the changes in atomic properties.\\3. Overemphasis on searching: While searching for relevant Wikipedia articles is a good idea, it's not enough to solve the problem. The output could benefit from a more in-depth analysis of the Rydberg equation and its applications to different atoms.\\4. Unclear role: Agent 3FeF claims to be a "Knowledgeable Expert" but doesn't demonstrate any expert-level knowledge or critical thinking in their output.\\ \\To improve the analysis, Agent 3FeF could provide a more comprehensive approach, such as:\\ \\1. Identifying specific aspects of the Rydberg equation that might not be applicable to the mentioned options (e.g., the assumption of a single electron or a specific type of atomic orbital)\\2. Providing examples or analogies to illustrate why the Rydberg equation might not work for certain options\\3. Offering alternative solutions or approaches to understanding the emission spectra of the mentioned atoms\\4. Explicitly stating the limitations and assumptions involved in applying the Rydberg equation to different atoms\\ \\By addressing these issues, Agent 3FeF can enhance the quality of their analysis and provide a more robust solution to the problem.
}
\end{tcolorbox}
\end{table*}

\begin{table*}[t]
\begin{tcolorbox}[notitle, colback=white, 
       boxrule=3pt, boxsep=0.5pt, enhanced, colframe=softpink,
       shadow={3pt}{-3pt}{0pt}{opacity=1,mygrey},
       title={\includegraphics[height=22pt]{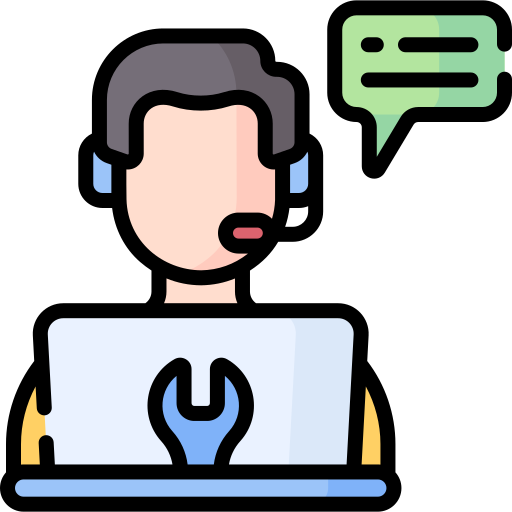} \large Mathematician 32fy: Round 1 (4/12)}]
\texttt{Thank you for the feedback from Agent 6PWE. As a mathematician, I must agree that my previous output was lacking in depth and analysis. Here's my revised answer:\\\\The Rydberg equation, $v = R_H(1/n_1^2 - 1/n_2^2)$, accurately predicts the UV-visible emission spectrum of the hydrogen atom because it assumes a single electron in a hydrogen-like atom. However, when applying this equation to other atoms, such as hydride ion (H-), deuterium atom (D), tritium atom (T), and helium cation (He+), we must consider the following limitations:\\\\1. The Rydberg equation assumes a single electron, which is not the case for the hydride ion (H-), which has two electrons. This means that the equation cannot accurately predict the emission spectrum of the hydride ion.\\2. Deuterium (D) and tritium (T) are isotopes of hydrogen, which have slightly different atomic properties. While the Rydberg equation can still be applied to these atoms, the predicted emission spectra will be slightly different due to the changes in atomic properties.\\\\On the other hand, the Rydberg equation can be applied to the helium cation (He+) because it assumes a single electron, and helium is a noble gas that has a single electron in its outermost orbital.\\\\In conclusion, the Rydberg equation is not applicable to the hydride ion (H-) because of its assumption of a single electron. The equation can be applied to deuterium (D) and tritium (T) with minor modifications to account for their atomic properties, but the predicted emission spectra will be slightly different.\\\\Thank you again for the feedback from Agent 6PWE. I hope this revised answer addresses the issues raised and provides a more comprehensive analysis of the Rydberg equation and its applications to different atoms.
}
\end{tcolorbox}
\end{table*}

\begin{table*}[t]
\begin{tcolorbox}[notitle, colback=white, colframe=softpurple,
       boxrule=3pt, boxsep=0.5pt, enhanced, 
       shadow={3pt}{-3pt}{0pt}{opacity=1,mygrey},
       title={\includegraphics[height=22pt]{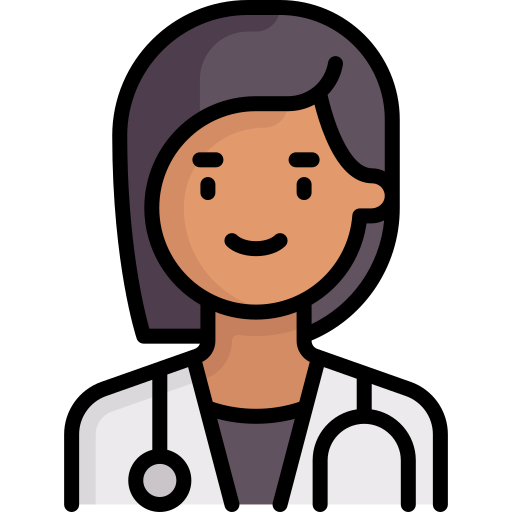} \large Psychologist vtXr: Round 1 (5/12)}]
\texttt{What a delightful exercise in critical thinking and problem-solving! As a psychologist, I'd like to offer some insights from a psychological perspective.\\\\Firstly, I'd like to commend Agent 3FeF for acknowledging the limitations of their initial output and recognizing the need for a more comprehensive approach. This is a crucial aspect of critical thinking, and it's excellent that they're willing to learn from their mistakes.\\\\Secondly, Agent 32fy has provided an outstanding revised analysis that addresses the issues raised by Agent 6PWE. Their output is now more detailed, specific, and evidence-based, demonstrating a clear understanding of the Rydberg equation and its limitations. The examples they provided to illustrate the applicability of the equation to different atoms are excellent, and the conclusion is well-supported by the logic of the analysis.\\\\From a psychological perspective, I'd like to highlight the importance of self-awareness and humility in critical thinking. By acknowledging their own limitations and being open to feedback, both Agents 3FeF and 32fy have demonstrated a high level of self-awareness. Furthermore, Agent 32fy's willingness to revise their output and provide a more comprehensive analysis shows an excellent grasp of the concept of constructive feedback.\\\\In conclusion, as a psychologist, I'm particularly pleased to see such a high level of critical thinking and problem-solving skills on display. I hope that these agents will continue to refine their skills and apply them to even more complex problems in the future.
}
\end{tcolorbox}
\end{table*}

\begin{table*}[t]
\begin{tcolorbox}[notitle, colback=white, colframe=softbrown,
       boxrule=3pt, boxsep=0.5pt, enhanced, 
       shadow={3pt}{-3pt}{0pt}{opacity=1,mygrey},
       title={\includegraphics[height=22pt]{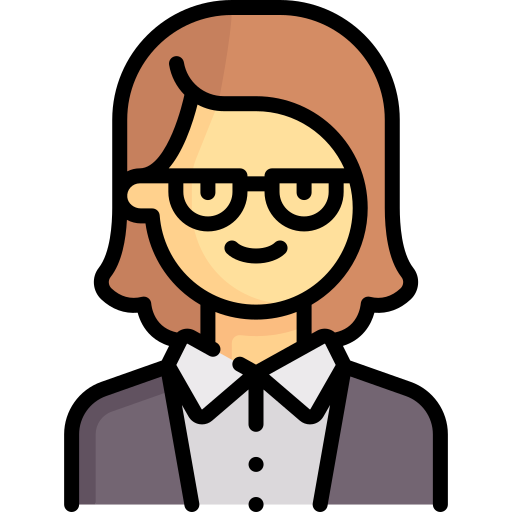} \large Historian 53AY: Round 1 (6/12) \& \includegraphics[height=22pt]{figure/agent3.png} \large Knowlegable Expert 3FeF: Round 2 (7/12)}]
\texttt{None.
}
\end{tcolorbox}
\end{table*}

\begin{table*}[t]
\begin{tcolorbox}[notitle, colback=white, colframe=softgreen,
       boxrule=3pt, boxsep=0.5pt, enhanced, 
       shadow={3pt}{-3pt}{0pt}{opacity=1,mygrey},
       title={\includegraphics[height=22pt]{figure/agent1.png} \large Critic 6PWE: Round 2 (8/12)}]
\texttt{Let's review the outputs of the other agents and identify potential issues:\\\\Agent 53AY:\\\\1. The initial output is correct that the Rydberg equation predicts the UV-visible emission spectrum of the hydrogen atom.\\2. However, the reason provided for why the Rydberg equation is not applicable to the hydride ion (H-) is limited. The equation is not only affected by the number of electrons but also by the atomic properties of the atom, such as the mass of the nucleus and the electron configuration.\\3. The output does not consider the possibility of applying the Rydberg equation to deuterium, tritium, and helium cation with some modifications.\\\\Agent 3FeF:\\\\1. The output is incomplete and lacks specific analysis. It merely states that the entities should be searched in Wikipedia without providing any relevant information or insights.\\\\Agent 32fy:\\\\1. The initial output is lacking in depth and analysis, as acknowledged by the agent themselves.\\2. The revised output provides a more detailed analysis, but it still has some limitations:\\+ It implies that the Rydberg equation can't be applied to deuterium and tritium at all, which is not entirely accurate. With some modifications, the equation can still be applied to these isotopes.\\+ The conclusion that the Rydberg equation is not applicable to the hydride ion (H-) due to its assumption of a single electron is correct, but the explanation could be more detailed.\\+ The analysis could be more comprehensive by considering the atomic properties of helium cation (He+) and why the Rydberg equation is applicable to it.\\\\In summary, while the agents have made some efforts to analyze the topic, they still lack a thorough understanding of the Rydberg equation and its limitations. A more in-depth analysis would consider the atomic properties of the atoms, the modifications required for deuterium, tritium, and helium cation, and the resulting differences in the predicted emission spectra.
}
\end{tcolorbox}
\end{table*}

\begin{table*}[t]
\begin{tcolorbox}[notitle, colback=white, colframe=softpink,
       boxrule=3pt, boxsep=0.5pt, enhanced, 
       shadow={3pt}{-3pt}{0pt}{opacity=1,mygrey},
       title={\includegraphics[height=22pt]{figure/agent4.png} \large Mathematician 32fy: Round 2 (9/12)}]
\texttt{I understand that the task is to identify which option is incorrect with regards to the applicability of the Rydberg equation to predict UV-visible emission spectrum.\\\\After reviewing the outputs of other agents, I agree with Agent 6PWE that Agent 3FeF's output lacks analysis and specific information, merely suggesting searching for Wikipedia articles without applying the concepts to the problem.\\\\Based on the Rydberg equation, $v = R_H(1/n_1^2 - 1/n_2^2)$, and the atomic properties of the atoms, I believe that the Rydberg equation is not applicable to the hydride ion (H-). This is because the equation assumes a single electron, which is not the case for negatively charged ions like H-.\\\\nHowever, the Rydberg equation can be applied to deuterium, tritium, and helium cation with some modifications, considering the changes in atomic properties. This is because these atoms have different electron configurations and masses of the nucleus.\\\\Therefore, the correct answer is Option A: hydride ion, H-.\\\\As for Agent 32fy's revised output, I agree that it is more comprehensive and detailed than Agent 3FeF's output. However, I would like to point out that the analysis could still be improved by considering the atomic properties of helium cation (He+) and why the Rydberg equation is applicable to it.\\\\Overall, I believe that my analysis takes into account the limitations of the Rydberg equation and provides a more comprehensive understanding of its applicability to different atoms.
}
\end{tcolorbox}
\end{table*}

\begin{table*}[t]
\begin{tcolorbox}[notitle, colback=white, colframe=softpurple,
       boxrule=3pt, boxsep=0.5pt, enhanced, 
       shadow={3pt}{-3pt}{0pt}{opacity=1,mygrey},
       title={\includegraphics[height=22pt]{figure/agent2.png} \large Psychologist vtXr: Round 2 (10/12)}]
\texttt{Thank you for sharing the outputs of the other agents. As a psychologist, I'd like to offer a more nuanced and scientific approach to this problem.\\\\The Rydberg equation, $v = R_H(1/n_1^2 - 1/n_2^2)$, is a fundamental concept in atomic physics. It's not just a simple formula, but rather a mathematical expression that describes the energy levels of an atom. The equation assumes that the atom has a single electron, which is a simplifying assumption.\\\\Considering the atomic properties of the entities mentioned, we can analyze each option:\\\\Option A: Hydride ion (H-). As mentioned, the Rydberg equation assumes a single electron, which is not the case for the hydride ion, as it has two electrons. Therefore, the equation is not directly applicable to the hydride ion.\\\\Option B: Deuterium atom (D). Deuterium is an isotope of hydrogen, and its atomic mass and electron configuration are very similar to those of the hydrogen atom. With some modifications, the Rydberg equation can still be applied to deuterium, but the predicted emission spectrum would be slightly different due to the increased mass of the nucleus.\\\\Option C: Tritium atom (T). Tritium is another isotope of hydrogen, and its properties are also similar to those of the hydrogen atom. Again, with some modifications, the Rydberg equation can be applied to tritium, but the predicted emission spectrum would be slightly different due to the increased mass of the nucleus.\\\\Option D: Helium cation (He+). The Rydberg equation can be applied to the helium cation, as it assumes a single electron, which is the case for the helium cation.\\\\In conclusion, while the Rydberg equation is applicable to all the options, the hydride ion (H-) is the only one where the equation is not directly applicable due to the assumption of a single electron. The other options - deuterium, tritium, and helium cation - can be analyzed using modified versions of the Rydberg equation that take into account the atomic properties of the atoms.\\\\I hope this analysis provides a more comprehensive understanding of the Rydberg equation and its limitations.
}
\end{tcolorbox}
\end{table*}

\begin{table*}[t]
\begin{tcolorbox}[notitle, colback=white, colframe=softbrown,
       boxrule=3pt, boxsep=0.5pt, enhanced, 
       shadow={3pt}{-3pt}{0pt}{opacity=1,mygrey},
       title={\includegraphics[height=22pt]{figure/agent6.png} \large Historian 53AY: Round 2 (11/12) \& \large Final Decision:  (12/12)}]
\texttt{The Rydberg equation, $v = R_H(1/n_1^2 - 1/n_2^2)$, accurately predicts the UV-visible emission spectrum of the hydrogen atom. A form of the Rydberg equation may also be used to predict the UV-visible emission for all of the following EXCEPT option A: hydride ion, H-.\\\\The reason for this exception is that the hydride ion (H-) has two electrons, whereas the Rydberg equation assumes a single electron. This means that the equation cannot accurately predict the emission spectrum of the hydride ion. \\
\rule[0mm]{\linewidth}{0.5pt}
A \textcolor{green(pigment)}{\Checkmark}
}
\end{tcolorbox}
\end{table*}

\end{document}